\documentclass[5p,times,procedia]{elsarticle}
\usepackage{ecrc}
\usepackage[nolist]{acronym}
\usepackage{comment}
\usepackage{footmisc}
\usepackage{fontawesome5}

\acrodef{bansai}[BANSAI]{\textbf{B}ridging the AI \textbf{A}doption Gap via \textbf{N}euro\textbf{s}ymbolic \textbf{AI}}
\acrodef{sme}[SME]{small- or medium-sized enterprise}
\acrodef{trl}[TRL]{technology readiness level}
\acrodef{ml}[ML]{machine learning}
\acrodef{ai}[AI]{artificial intelligence}
\acrodef{dl}[DL]{deep learning}
\acrodef{rl}[RL]{reinforcement learning}
\acrodef{dnn}[DNN]{deep neural network}
\acrodef{dp}[$\partial$P]{differentiable programming}
\acrodef{vr}[VR]{virtual reality}
\acrodef{mp}[MP]{movement primitive}
\acrodef{plm}[PLM]{product lifecycle management}
\acrodef{mes}[MES]{manufacturing excution system}
\acrodef{tamp}[TAMP]{task and motion planning}
\acrodef{krr}[KR\&R]{knowledge representation and reasoning}
\acrodef{ide}[IDE]{integrated development environment}
\acrodef{dcg}[DCG]{differentiable computation graph}

\volume{00}

\firstpage{1}

\journalname{Procedia CIRP}

\runauth{Benjamin Alt et al.}

\jid{trpro}

\usepackage{amssymb}

\usepackage[figuresright]{rotating}

\usepackage[bookmarks=false]{hyperref}
    \hypersetup{colorlinks,
      linkcolor=blue,
      citecolor=blue,
      urlcolor=blue}

\begin{document}
\begin{frontmatter}



\dochead{57th CIRP Conference on Manufacturing Systems 2024 (CMS 2024)}%

\title{BANSAI: Towards Bridging the AI Adoption Gap in Industrial Robotics with Neurosymbolic Programming}

\author[a,b]{Benjamin Alt\textsuperscript{*,}} 
\author[c]{Julia Dvorak}
\author[a]{Darko Katic}
\author[a]{Rainer Jäkel}
\author[b]{Michael Beetz}
\author[c]{Gisela Lanza}

\address[a]{ArtiMinds Robotics, Albert-Nestler-Str. 11, 76131 Karlsruhe, Germany}
\address[b]{Institute for Artificial Intelligence, University of Bremen, Am Fallturm 1, 28359 Bremen}
\address[c]{wbk Institute of Production Science, Karlsruhe Institute of Technology, Gotthard-Franz-Straße 5, 76131 Karlsruhe}

\aucores{* Corresponding author. Tel.: +49 721 509998-66. {\it E-mail address:} benjamin.alt@uni-bremen.de}

\begin{abstract}
Over the past decade, deep learning helped solve manipulation problems across all domains of robotics. At the same time, industrial robots continue to be programmed overwhelmingly using traditional program representations and interfaces. This paper undertakes an analysis of this “AI adoption gap” from an industry practitioner’s perspective. In response, we propose the \acs{bansai} approach (\textbf{B}ridging the AI \textbf{A}doption Gap via \textbf{N}euro\textbf{s}ymbolic \textbf{AI}). It systematically leverages principles of neurosymbolic AI to establish data-driven, subsymbolic program synthesis and optimization in modern industrial robot programming workflow. BANSAI conceptually unites several lines of prior research and proposes a path toward practical, real-world validation.
\end{abstract}


\end{frontmatter}



\section{Introduction}
\label{sec:introduction}

Deep neural networks and subsymbolic learning have progressed tremendously over the past decade, producing increasingly promising results in the domain of program synthesis and robot control \cite{kroemer_review_2021}. While the use of robots in the manufacturing industries is ubiquitous, the current degree of industry adoption of \acl{ai}-based robot program synthesis and optimization remains very limited, particularly with regard to \ac{dl} \cite{heimann_industrial_2020}. This reflects a broader phenomenon in the manufacturing industry, where \ac{ai} adoption lags behind the academic state of the art, with a ``lack of substantial evidence of industrial success'' at \acp{trl} 5 and beyond \cite{peres_industrial_2020}. The lack of \ac{ai} adoption for robot programming stands in stark contrast to perception tasks such as visual inspection, object recognition or anomaly detection, where \ac{ai} systems have found widespread acceptance \cite{sunderhauf_limits_2018}. Facing rising prevalence of high-mix, low-volume applications, bridging this ``\ac{ai} adoption gap'' can greatly reduce robot programming overhead and make robotic automation viable for use cases requiring frequent reprogramming or reparameterization.

In this paper, we propose that neurosymbolic programming - a principled combination of symbolic \ac{ai} and \acf{dl} for program representation, synthesis and optimization - can overcome this gap. We describe \acsu{bansai} (\acl{bansai}), an approach for the application of neurosymbolic programming to industrial robotics. To that end, we contribute an analysis of the \ac{ai} adoption gap,  highlighting a mismatch between the requirements imposed by the industrial robot programming and deployment process and the exigencies of state-of-the-art \ac{ai}-based manipulation, program synthesis and optimization approaches. We propose that the unique properties of neurosymbolic \ac{ai} can serve as the basis for \ac{ai} technologies which are fundamentally compatible with the real safety, performance and human-machine-interaction constraints faced by robot programmers and automation engineers. We then describe \acs{bansai}, a novel approach to robot programming designed specifically for real-world industrial application, which leverages neurosymbolic \ac{ai} to provide \ac{ai} assistance across the complete programming and deployment process.

\section{The \ac{ai} Adoption Gap in Industrial Robot Programming}
\label{sec:ai_adoption_gap}
The comparatively slow pace of adoption of \ac{ai} by industry practitioners has been noted both in a recent survey of industrial \ac{ai} \cite{peres_industrial_2020} as well as industry reports \cite{brosset_scaling_2019,winkler_accelerating_2019}. In recent surveys of industrial robot programming methods \cite{heimann_industrial_2020,krot_intuitive_2019}, \ac{ai}-based methods are not mentioned, despite considerable research activity. We provide an analysis of the reasons behind this ``\ac{ai} adoption gap'', as well as an approach to overcome it (see chapter \ref{sec:bansai}). Because upfront robot programming comprises only a small part of the effort involved in bringing a robotic workstation from conception to production, we consider the complete deployment process up to the final operation of the programmed robot within the larger production context.

\begin{figure*}
  \centering
  \includegraphics[width=\textwidth]{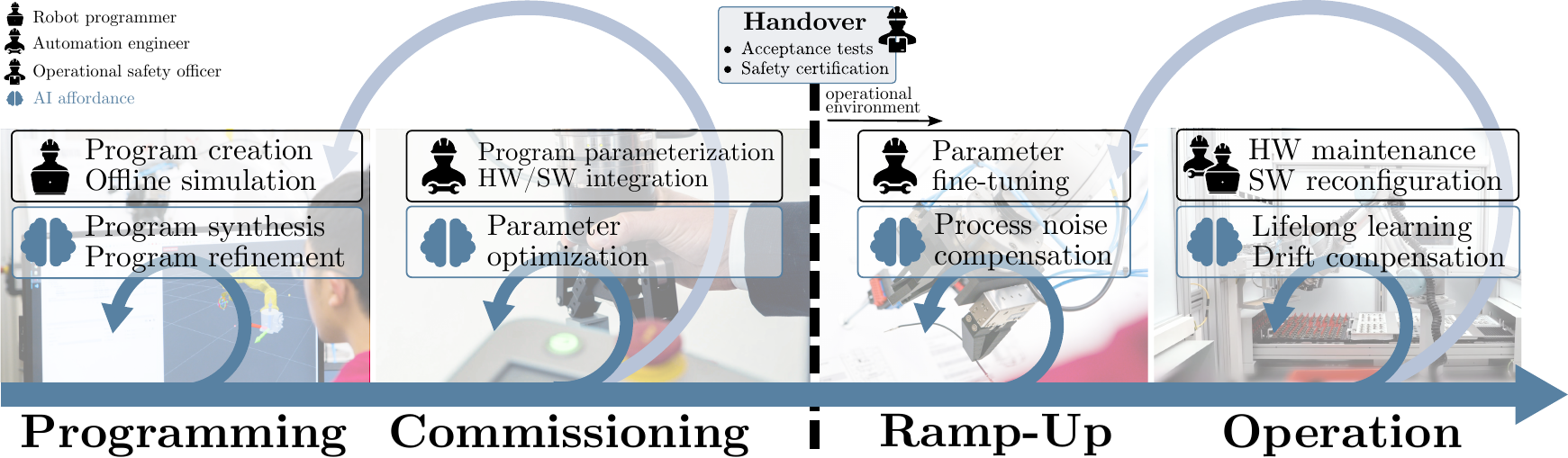}
  \caption{A simplified model of the industrial robot programming process, the roles and involvement of human actors, as well as opportunities for \ac{ai} assistance.}
  \label{fig:simplified_industrial_robot_programming_process}
\end{figure*}

\subsection{Industrial Robot Programming and Deployment}
\label{sec:robot_programming_workflow}
While the process of bringing an industrial robot workstation\footnote{We use the term ``workstation'' instead of ``workcell'' or ``robot cell'', as flexible human-robot co-workspaces become increasingly common.} from conception to operation can vary widely between companies, most follow variations of the robot programming and deployment process illustrated in Fig. \ref{fig:simplified_industrial_robot_programming_process}:

\begin{enumerate}
  \item \textbf{Programming.} The initial robot program is created by the robot programmer. This requires considerable domain expertise, particularly for complex manipulation tasks involving force-dynamic interaction with workpieces (e.g. tight insertion, cable manipulation or sanding).
  \item \textbf{Commissioning.} The robot software is deployed in the physical robot workstation. It is iteratively refined until requirements with respect to cycle time, robustness and quality are met. The refinement of program parameters is time-consuming and requires a high degree of expertise.
  \item \textbf{Handover.} Most robot workstations are commissioned offsite by systems integrators or in-house engineers. Acceptance testing and safety certification typically occur either directly before or soon after integration of the workstation into the factory. For many large companies such as automotive manufacturers, robot programs must additionally comply with formal internal company standards. 
  \item \textbf{Ramp-Up.} After certification, the workstation is integrated into the assembly line and the production velocity is incrementally increased until the final production cycle times and robustness are achieved. Ramp-up is characterized by repeated optimization of program parameters to compensate process noise (different suppliers, lighting conditions, vibrations, ...). Changes to program structure are infrequent, as the robot workstation may already have passed safety certification.
  \item \textbf{Operation.} The robot workstation is used in production, possibly over very long timeframes. Central challenges during operation are the degradation of performance due to wear and tear; the need for re-parameterization after repairs; and adaptation to new product variants (e.g. for small-batch production).
\end{enumerate}

\subsection{AI Challenges in Industrial Robot Programming}
The described process of programming industrial robots has unique properties distinguishing it from other domains. In the following paragraphs, we highlight five such properties, which must inform the design of \ac{ai} systems for robot programming. Conversely, limited \ac{ai} adoption can be partially explained by challenges posed by these properties.

\paragraph{High program complexity} Typical industrial robot programs span thousands of lines comprising varied motion and manipulation skills \cite{heimann_industrial_2020}. Symbolic program synthesis approaches can handle structural complexity but lack expression for subsymbolic skill optimization. End-to-end \ac{dl} learns complex skills but does not scale to long sequences, particularly with \ac{rl} \cite{arents_smart_2022}. Moreover, data issues such as scarcity and drift severely limit \ac{dl} in production \cite{peres_industrial_2020}.

\paragraph{Heterogeneous execution environments} Industrial robots are typically embedded into a complex digital production infrastructure spanning \ac{plm}, \acp{mes} and process control systems, requiring complex communication and synchronization logic. Therefore, ``robust integration with legacy IT systems (such as ERP, PLM and MES applications) should be addressed proactively'' \cite{peres_industrial_2020}.

\paragraph{Real-world physical manipulation} Industrial robot programs aim to cause effects in the physical world, subject to sensor and process noise. While purely \ac{dl}-based approaches excel at perception and planning problems in observable discrete spaces, reliably solving real-world manipulation remains challenging \cite{sunderhauf_limits_2018,kroemer_review_2021}. As it requires exploratory executions during training, \ac{rl} is difficult for industrial contexts \cite{sunderhauf_limits_2018,arents_smart_2022}. Likewise, human demonstrations are impossible for many industrial tasks exceeding human strength or precision.

\paragraph{Human involvement} Industrial robot programming currently involves cycles of re-programming and re-parameterization by experts. However, \ac{dl} models' input-output relationships are typically not interpretable by humans \cite{peres_industrial_2020}. This "intelligibility" aspect of explainability \cite{weld_challenge_2019} has been identified as crucial for \ac{ai}-assisted programming \cite{adadi_peeking_2018} - it is doubly crucial in industrial contexts where human expertise remains core \cite{sanneman_state_2021}. The requirement of human-editability calls for modular rather than end-to-end approaches, allowing engineers to modify parts of programs \cite{sunderhauf_limits_2018}.

\paragraph{High trust requirements} 
Beyond intelligibility for human interaction, industrial application demand \textit{trustworthy} programs able to afford \textit{explanation} and \textit{certification} \cite{huang_survey_2020}. Certification requires robot programs to be able to make hard guarantees about their behavior. However, deep networks can be highly sensitive to small perturbations in their inputs \cite{bruna_intriguing_2013}, and formal verification does not scale beyond small networks \cite{huang_survey_2020}. Low interpretability entails lack of perceived trust and reliability \cite{peres_industrial_2020}, hindering adoption. Conversely, explainable systems are more likely adopted by industry practitioners \cite{weld_challenge_2019}.

\section{Neurosymbolic Robot Programming}
\label{sec:neurosymbolic_robot_programming}
In recent years, \textit{neurosymbolic \ac{ai}} has received greatly increased attention \cite{hitzler_neuro-symbolic_2022-1,sarker_neuro-symbolic_2022}. Neurosymbolic \ac{ai} combines symbolic \ac{ai} methods such as \ac{ai} planning, knowledge bases and symbolic reasoning with subsymbolic, neural representations and algorithms such as \acp{dnn} and backpropagation. A \textit{neurosymbolic program} is ``a program that uses neural components and either symbolic components or symbolic compositions'' \cite{chaudhuri_neurosymbolic_2021}: Examples include hybrid program representations in which some computations are realized by neural networks but where control flow or I/O is handled by symbolic primitives \cite{innes_differentiable_2019,alt_robot_2021}; hierarchical neural architectures \cite{valkov_houdini_2018,frans_meta_2018}; or subsymbolic learning algorithms synthesizing symbolic programs \cite{verma_imitation-projected_2019,shah_learning_2020}. In the context of robot programming, state-of-the-art neurosymbolic approaches typically represent robot programs as graphs of modules with well-defined, documented behavior and interfaces, where the modules themselves are (partially) subsymbolic \cite{seker_conditional_2019,kulak_active_2021}. This allows for module reuse, explainability and use of intuitive user interfaces for manual programming at the symbolic (structural) level, while retaining most advantages of neural architectures such as learnability and partial model-freeness at the module level.

We propose that neurosymbolic programming combines the benefits of symbolic and subsymbolic \ac{ai} in a way which makes it uniquely suited for industrial robot programming: By virtue of their reliance on symbolic composition, neurosymbolic program representations are inherently modular \cite{chaudhuri_neurosymbolic_2021}, allowing to leverage the scalability of symbolic planners to the highly complex program structures typical for industrial appliations.

Symbolic composition further permits the use of symbolic \ac{krr} systems for program synthesis \cite{sarker_neuro-symbolic_2022,alt_knowledge-driven_2023,kazhoyan_towards_2020}. Symbolic knowledge representations can efficiently encode existing domain knowledge of e.g. assembly-line workers and robot programmers, without requiring the conversion of this knowledge into training data for a neural network. Crucially for practical applications, symbolic composition enables the re-use of algorithmic knowledge embedded in existing planners \cite{shah_learning_2020}.

An additional consequence of symbolic composition is the intellegibility of neurosymbolic programs at the structural level \cite{chaudhuri_neurosymbolic_2021}. Symbolic composition requires neural program components to be hidden behind well-defined interfaces, permitting human programmers to symbolically compose complex programs from encapsulated neural primitives without requiring \ac{dl} expertise. Moreover, it enables the gradual replacement of traditional program components by neural components without disrupting the overall programming and deployment process.

Lastly, symbolic composition enables the mixture of neural and symbolic program components within a hybrid program representation. Such representations greatly facilitate the integration of learnable components with the I/O and synchronization ``glue code'' required to integrate industrial robot programs into the larger factory context.

\section{BANSAI: A Neurosymbolic Approach to Industrial Robot Programming}
\label{sec:bansai}
We propose \acs{bansai} (\acl{bansai}), an approach for practical \ac{ai}-assisted industrial robot programming using principles of neurosymbolic learning and inference. At its core, \ac{bansai} follows a ``bottom-up'' philosophy \cite{sanneman_state_2021}: \ac{ai} solutions and workflows should be designed to fit the needs of the application and context in which they are employed. Consequently, \ac{bansai} (a) reflects the robot programming and deployment process (see Fig. \ref{fig:simplified_industrial_robot_programming_process}) as it is practiced across the manufacturing industry, with the aim of allowing a gradual introduction of \ac{ai} assistance without disturbing the overall process; and (b) uses neurosymbolic \ac{ai} to address the \ac{ai} challenges posed by industrial robot programming and deployment, with the aim of facilitating its adoption by practitioners and decisionmakers. \ac{bansai} unifies prior work by the authors \cite{alt_robot_2021,alt_method_2022,alt_heuristic-free_2022,alt_knowledge-driven_2023,alt_robogrind_2024} into a coherent neurosymbolic robot programming approach and proposes a concrete workflow for the \ac{ai}-assisted programming of industrial robots.

\begin{figure*}[t]
  \centering
  \includegraphics[width=\textwidth]{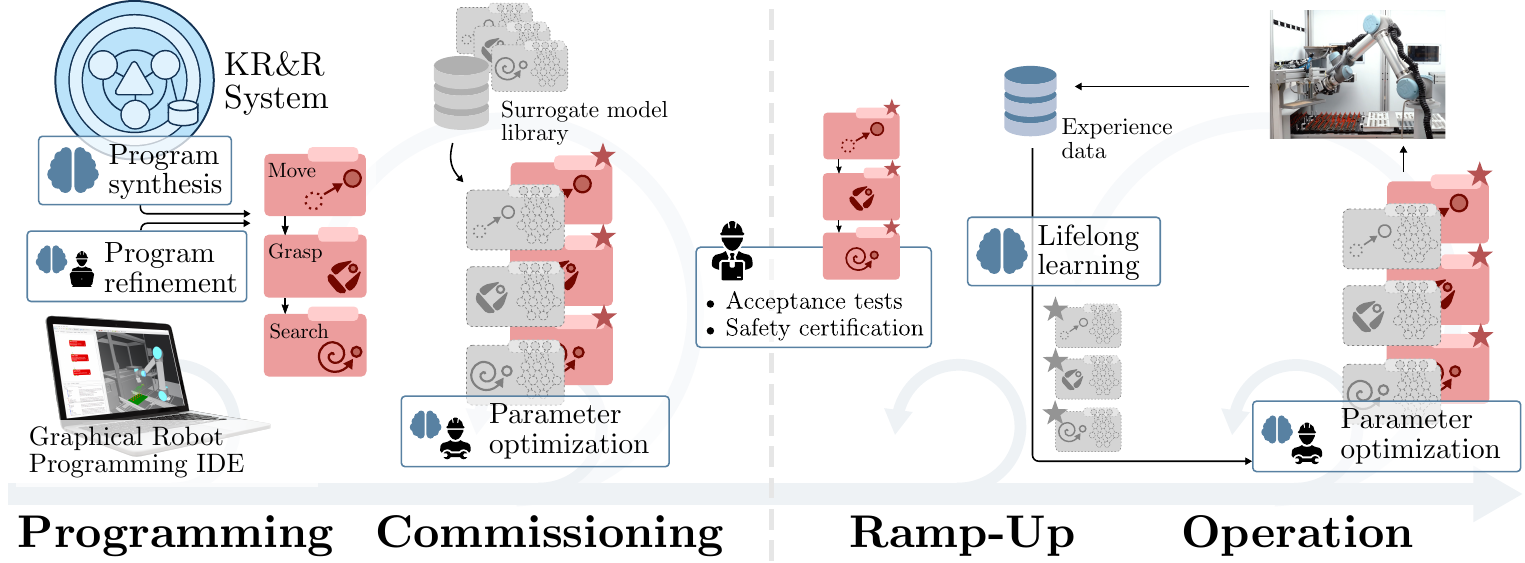}
  \caption{The \ac{bansai} workflow for AI-assisted industrial robot programming. The use of a dual symbolic-subsymbolic program representation (red/grey) enables the seamless integration of \ac{ai} assistance ({\scriptsize \faIcon{brain}}) into typical industrial robot programming processes.}
  \label{fig:ai_assisted_industrial_robot_programming_workflow}
\end{figure*}

\subsection{Neurosymbolic Programming with a Dual Program Representation}

The technical foundation of \ac{bansai} is a dual symbolic-subsymbolic representation of robot programs \cite{alt_robot_2021,alt_method_2022}. It combines a traditional, skill-based robot program representation for user interaction, motion planning and robot control with a neural ``surrogate'' representation of the same program for learning and parameter optimization. 

\paragraph{Graphical robot programming} The symbolic component of the program representation is a traditional, skill-based representation, which is used for execution on the robot as well as for interaction with human users (see Fig. \ref{fig:ai_assisted_industrial_robot_programming_workflow} (red)). Representing programs as graphs of primitive skills with well-defined and documented behavior allows human programming experts to use intuitive interfaces to create, modify or read programs and fosters trust in both the programming system and the programs themselves. Moreover, industrial skill-based program representations (e.g. ArtiMinds ARTM \cite{schmidt-rohr_artiminds_2013}, Universal Robots PolyScope\cite{universal_robots_polyscope_2018}) and most skill frameworks proposed by researchers (e.g. DMPs \cite{schaal_dynamic_2006}, ProMPs \cite{paraschos_probabilistic_2013-1}) allow establishing guarantees about the behavior of the robot at runtime, making certification a possibility. \ac{bansai} does not impose any specific constraints on the implementation of the skills (and their combination into programs) themselves: The only requirement is that skills provide a degree of explainability and allow for symbolic composition \cite{alt_robot_2021}.

\paragraph{Learning \& optimization via neural surrogates} To enable \ac{ai}-based program synthesis and optimization, the dual program representation proposes a neural surrogate (see Fig. \ref{fig:ai_assisted_industrial_robot_programming_workflow} (grey)) to the symbolic robot program \cite{alt_robot_2021}. Neural surrogates are neural networks which are trained to approximate a system, and are then used as surrogates for the system in downstream tasks \cite{holena_neural_2010}. \ac{bansai} proposes to use learned neural surrogates (``surrogate models'') of robot skills to optimize the original skills' parameters, e.g. with respect to different workpieces, changing environments or to compensate for long-horizon drifts. To that end, a library of neural surrogates, one for each available robot skill, is trained (on simulated data) offline. For a given graphical robot program, the corresponding graph of neural surrogate models (``surrogate program'', a \ac{dcg}) can be constructed automatically. During commissioning, the surrogate program is trained to approximate the behavior of the robot program it represents. The learned surrogate program can then be used to optimize the original program's parameters via a gradient-based optimizer. For further details, we refer to prior work by the authors \cite{alt_robot_2021}, which provides a detailed description of the algorithm and a comprehensive evaluation for multiple different symbolic skill frameworks, robots and application scenarios.

\paragraph{\ac{krr}-based metaprogramming} A corollary of using a dual symbolic-subsymbolic program representation is that it affords symbolic composition, which in turn enables the use of symbolic \ac{krr} systems for program synthesis. \ac{bansai} proposes to realize program synthesis in the form of \ac{krr}-driven metaprogramming: To encode domain and process knowledge in a semantic knowledge base, which, along with a set of general inference rules (metaprograms), permits the bootstrapping of complex robot (sub-)programs to solve tasks in a variety of domains. Program synthesis via symbolic \ac{krr} is inherently explainable, as it is always possible to enumerate the facts in the knowledge base which made an inference query true or false. Moreover, it permits the efficient use of existing process and domain knowledge. In prior work \cite{alt_knowledge-driven_2023}, we have proposed a \ac{krr}-based metaprogramming system using  KnowRob \cite{beetz_knowrob_2018} and the ARTM industrial robot program representation \cite{schmidt-rohr_artiminds_2013}. The proposed system has been evaluated in retail fetch-and-place \cite{alt_knowledge-driven_2023} as well as industrial surface treatment applications \cite{alt_robogrind_2024}.

\subsection{The \ac{bansai} Workflow}
\label{sec:programming_wf}
One core intuition behind the \ac{bansai} approach is that \ac{ai} assistance functions must seamlessly integrate into the industrial robot programming and deployment processes used in practice. Our proposed workflow for \ac{ai}-assisted industrial robot programming is shown in figure \ref{fig:ai_assisted_industrial_robot_programming_workflow}, though the flexibility of \ac{bansai} ensures its applicability to other, domain- or company-specific variants of this process.

\paragraph{Programming}
The initial robot program is created automatically via the \ac{krr}-metaprogramming \cite{alt_knowledge-driven_2023,alt_robogrind_2024}, given a high-level description or demonstration of the task by a human expert. The generated program is a skill-based robot program in an established industrial robot program representation, allowing robot programmers to refine it as needed using graphical tools and offline simulators \cite{schmidt-rohr_artiminds_2013}.

\paragraph{Commissioning}
During commissioning, the equivalent surrogate program to the robot program can be created automatically and fine-tuned in an unsupervised manner on data collected passively over the course of the commissioning process \cite{alt_robot_2021}. The parameters of the robot skills are optimized using a gradient-based optimizer over the surrogate program. For the robot programmer, the time-consuming trial-and-error of parameter tweaking is reduced to specifying a loss function for the automatic optimization, typically a function of the cycle time and robustness requirements.

\paragraph{Handover}
One of the core technical principles of \ac{bansai} is that the skill-based robot program, as opposed to its neural surrogate, is executed on the robot. For this reason, the handover process, including acceptance tests and safety certification, is not impacted, despite both the structure and parameters of the robot program were created and optimized using \ac{ai} systems.

\paragraph{Ramp-Up}
The ramp-up phase is characterized by iterative re-parameterization of the program until performance and robustness criteria are met in the operative environment. As during commissioning, gradient-based optimization over neural surrogate models can automate this parameter tweaking \cite{alt_robot_2021}. If the robot workstation has been safety-certified for a range of program parameters (e.g. robot velocities, forces or torques), \ac{ai}-based parameter optimization in these limits does not require re-certification.

\begin{figure*}[t]
  \centering
  \includegraphics[width=\textwidth]{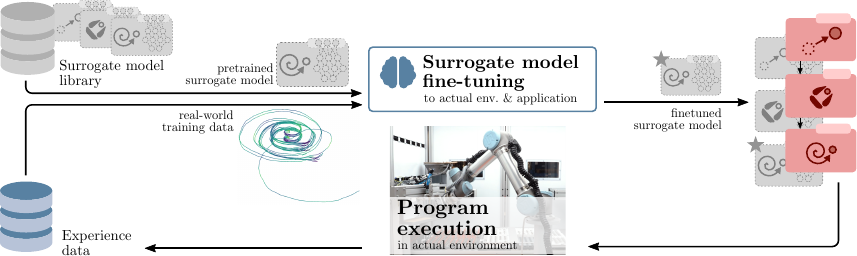}
  \caption{Lifelong learning in the \ac{bansai} workflow: Continuously fine-tuning surrogate models during ramp-up and production keeps the surrogate models updated for downstream parameter optimization.}
  \label{fig:rampup_workflow}
\end{figure*}

\paragraph{Operation}
During operation, challenges involve program reparameterization in response to drift caused by e.g. wear and tear, but also sudden  changes to the program parameterization in response to mechanical reconfiguration of the workstation or adaptation to new product variants. In \cite{alt_heuristic-free_2022}, we have shown that the surrogate model architecture proposed in \cite{alt_robot_2021} affords unsupervised lifelong learning on data gathered passively during operation, which keeps the surrogate program up-to-date with slow drifts or sudden shifts. This allows the proposed \ac{ai}-based parameter optimizer to constantly keep program parameters in the optimal range. Fig. \ref{fig:rampup_workflow} illustrates lifelong learning of surrogate models in the \ac{bansai} context.

\vspace{\baselineskip}

To our knowledge, \ac{bansai} is the first concept for \ac{ai}-assisted robot programming which respects the requirements and constraints of industrial applications. It is also the first \ac{ai}-based industrial robot programming concept to take a process-centric view, aiming to provide solutions to the programming challenges arising during the entire robot program lifecycle. Instead of tailoring the robot programming workflow around the requirements of an AI assistant, it leverages neurosymbolic \ac{ai} to integrate \ac{ai} asisstance functions into the existing robot programming process. The dual program representation enables the learning and gradient-based optimization afforded by neural architectures as well as the use of symbolic planners and reasoners. Reliance on the neural surrogate pattern ensures that the program executed on the robot is always explainable, human-editable and certifiable. While the  technology components of \ac{bansai} have been individually evaluated on real-world scenarios \cite{alt_robot_2021,alt_knowledge-driven_2023,alt_method_2022,alt_heuristic-free_2022,alt_robogrind_2024}, an implementation and evaluation of \ac{bansai} as a whole is currently being undertaken.

\section{Conclusion}
\label{sec:conclusion}
We have characterized the \ac{ai} adoption gap in industrial robot programming and proposed several \ac{ai} challenges posed by the robot programming process practiced in the manufacturing industry. Neurosymbolic programming combines symbolic and subsymbolic \ac{ai} in ways uniquely suited to address the particular requirements of industrial robot programming. Based on this insight, we presented \ac{bansai}, a neurosymbolic approach which addresses the specific challenges faced by robot programmers. Our insights highlight the importance of considering the needs of practitioners when designing \ac{ai} algorithms, particularly in applied disciplines such as industrial robotics. \ac{bansai} proposes an overarching workflow that combines state-of-the-art approaches from \ac{dl}-based program optimization \cite{alt_robot_2021,alt_method_2022,alt_heuristic-free_2022} and symbolic program synthesis \citep{alt_knowledge-driven_2023,alt_robogrind_2024} to realize highly flexible workflows, where some functionality is realized autonomously by \ac{ai}, while enabling intuitive human involvement where beneficial. Future work will focus on the implementation of a unified software framework and user interface for neurosymbolic robot programming, and the evaluation of the overall approach on a real-world production scenario.


\section*{Acknowledgements}

This work was supported by the German Federal Ministry of Education and Research (BMBF) grants 02L19C255 and 01DR19001B.

\bibliography{isrr_2022_bibliography}
\bibliographystyle{elsarticle-num}


\clearpage\onecolumn

\end{document}